\documentclass[letterpaper, 10 pt, conference]{ieeeconf}  %
\IEEEoverridecommandlockouts                              %
\usepackage{cite}
\usepackage{graphicx}
\usepackage{multirow}
\usepackage{amsmath}
\usepackage{amssymb}
\usepackage{amsfonts}
\usepackage{blindtext}
\usepackage{subcaption}
\usepackage[export]{adjustbox}
\usepackage{url}
\usepackage[colorinlistoftodos, german]{todonotes}
\usepackage[T1]{fontenc}
\usepackage[utf8x]{inputenc}
\usepackage[english]{babel}
\usepackage{hyperref}
\usepackage{booktabs, threeparttable}
\usepackage[absolute,showboxes]{textpos}
\usepackage{epsfig}
\usepackage{times}
\usepackage{mathtools}
\usepackage{mathrsfs}
\usepackage{siunitx}
\usepackage{multicol}
\usepackage[algoruled, noline, noend]{algorithm2e}
\usepackage{tikz}
\usepackage{float}
\usetikzlibrary{%
  calc,
  fit,
  shapes,
  backgrounds
}
\usepackage{pgfplots}
\usepackage{layouts}
\usepackage[switch]{lineno}
\usepackage{tabularx}
\usepackage{setspace}

\pgfplotsset{compat=newest}
\usepgfplotslibrary[groupplots]
\hypersetup{
    colorlinks=true,
    linkcolor=black,
    citecolor=black,
    filecolor=black,
    urlcolor=black,
pdfinfo={
Title={Efficient Sampling in POMDPs with Lipschitz Bandits for Motion Planning in Continuous Spaces},    
Author={Omer Sahin Tas},
Subject={Proceedings of IEEE Intelligent Vehicle Symposium}
}
}

\usepackage{xcolor}
\definecolor{color_ucb}{RGB}{44,123,182}
\definecolor{color_ucbv}{RGB}{171,217,233}
\definecolor{color_pomcp}{RGB}{255,255,191}
\definecolor{color_poslb}{RGB}{215,25,28}
\definecolor{color_poslbv}{RGB}{253,174,97}
\definecolor{color_collision_term}{RGB}{77,175,74}
\definecolor{color_all_terms}{RGB}{55,126,184}
\definecolor{color_rough}{RGB}{141,160,203}
\definecolor{color_dense}{RGB}{252,141,98}

\usetikzlibrary{shapes.misc}
\tikzset{cross/.style={cross out, draw=black, minimum size=2*(#1-\pgflinewidth), inner sep=0pt, outer sep=0pt}, cross/.default={1pt}}

\newcommand{\ie}{\emph{i.e. }}

\newcommand{\abs}[1]{\left\lvert#1\right\rvert}

\newcommand{\collisiontwo}{\texttt{S\textsubscript{Coll}}}
\newcommand{\mczero}{\texttt{S\textsubscript{I-Lo}}}
\newcommand{\mctwo}{\texttt{S\textsubscript{I-Hi}}}
\newcommand{\stdevmae}{$\sigma_{\text{MAE}}$}

\usepackage{xspace}

\makeatletter
\DeclareRobustCommand\onedot{\futurelet\@let@token\@onedot}
\def\@onedot{\ifx\@let@token.\else.\null\fi\xspace}

\def\ie{i.e\onedot}

\def\cf{cf\onedot}

\def\etal{et al\onedot}
\makeatother

\newcommand{\mab}{\mbox{MAB}\xspace}
\newcommand{\ucb}{\mbox{UCB}\xspace}
\newcommand{\ucbv}{\mbox{UCB-V}\xspace}
\newcommand{\poslb}{\mbox{POSLB}\xspace}
\newcommand{\poslbv}{\mbox{POSLB-V}\xspace}
\newcommand{\pomdp}{\mbox{POMDP}\xspace}
\newcommand{\pomdps}{\mbox{POMDPs}\xspace}
\newcommand{\pomcp}{\mbox{POMCP}\xspace}
\newcommand{\tapir}{\mbox{TAPIR}\xspace}

\newcommand{\acc}[1]{\SI{#1}{\metre\per\second\squared}}

\newcommand{\optimum}[1]{{#1}^{\ast}}
\newcommand{\estimator}[1]{\hat{#1}}
\newcommand{\complexity}{O} %
\newcommand{\maximum}[1]{\underset{#1}{\operatorname{max}}}
\newcommand{\argmaximum}[1]{\operatorname{arg\,\underset{#1}{\operatorname{max}}}}
\newcommand{\minus}{\scalebox{0.6}{$-$}}

\newcommand{\lowerBound}[1]{{#1}^{\minus}}
\newcommand{\realNumbers}{\mathbb{R}}

\newcommand{\history}{{h}}
\newcommand{\samplingInterval}{{T_s}} %
\newcommand{\discountFactor}{{\gamma}}
\newcommand{\learningRate}{\eta}
\newcommand{\explorationConstant}{c}
\newcommand{\lipschitzConstant}{\mathcal{L}}
\newcommand{\state}{{s}}
\newcommand{\observation}{{o}}
\newcommand{\belief}{{b}}
\newcommand{\action}{{a}}
\newcommand{\actionSet}{{\mathcal{A}}}
\newcommand{\reward}{{r}}
\newcommand{\costScalingFactor}{{\zeta}}
\newcommand{\stateSpace}{{\mathbb{{S}}}}
\newcommand{\actionSpace}{{\mathbb{A}}}
\newcommand{\rewardSpace}{\mathbb{\MakeUppercase{\reward}}}
\newcommand{\observationSpace}{{\mathbb{O}}}
\newcommand{\rewardModel}{\mathcal{R}}
\newcommand{\qvalue}{{Q}}
\newcommand{\Value}{{V}}
\newcommand{\policy}{{\pi}}
\newcommand{\qvaluePolicy}[1]{\qvalue^{\policy_{#1}}}
\newcommand{\banditArm}{a}
\newcommand{\banditArmSet}{\mathcal{A}}
\newcommand{\banditRounds}{{T}}
\newcommand{\banditReward}{\mu}

\newcommand{\banditArmMeanReward}[3][\banditArm]{{#2{\banditReward}}_{#3} (#1)}

\newcommand{\positionCartesianX}{{x}}
\newcommand{\positionCartesianY}{{y}}
\newcommand{\positionFrenetLon}{{l}}  %

\newcommand{\speed}{{v}}
\newcommand{\speedReference}{{v_{\mathrm{ref}}}}

\newcommand{\acceleration}{{a}}  %

\newcommand{\decelerationMax}{\lowerBound{\acceleration}}
\newcommand{\accelerationComfort}{a_{\mathrm{cmf}}}
\newcommand{\decelerationComfort}{b_{\mathrm{cmf}}}
\newcommand{\jerk}{{j}}

\newcommand{\curvature}{{\kappa}}
\newcommand{\route}{\rho} %
\newcommand{\routeData}{p} %

\pgfmathdeclarefunction{lg10}{1}{ \pgfmathparse{ln(#1)/ln(10)}}

\TPMargin{5pt}

\newcommand{\copyrightstatement}{
    \begin{textblock}{0.84}(0.08,0.930)
         \noindent{
             \footnotesize{
                 \begin{spacing}{1.0}
                 \noindent \copyright 2021 IEEE.
                 Personal use of this material is permitted. Permission from IEEE must be obtained for all other uses, in any current or future media, including reprinting/republishing this material for advertising or promotional purposes, creating new collective works, for resale or redistribution to servers or lists, or reuse of any copyrighted component of this work in other works.\\[1mm]
                 \noindent
                 Accepted for publication in Proceedings of the IEEE Intelligent Vehicles Symposium (IV), Nagoya-Japan, 11-17 July 2021.
                \end{spacing}
            }
        }
    \end{textblock}
}

\title{\LARGE \bf
Efficient  Sampling  in  POMDPs  with  Lipschitz  Bandits \\for  Motion  Planning  in  Continuous  Spaces}
\author{
Ömer~Şahin~Taş,
Felix~Hauser,
and Martin~Lauer
\thanks{
Corresponding author: {\tt\small tas@fzi.de}. The authors are with
FZI Research Center for Information Technology and Karlsruhe Institute of Technology, 76131, \textsc{Germany}.
This paper is an extended version of our paper presented in Ö.\ Ş.\ Taş, F.\ Hauser, and M.\ Lauer, ``Efficient Sampling in POMDPs with Lipschitz Bandits for Motion Planning in Continuous Spaces,'' in \textit{Workshop on Planning, Perception, Navigation for Intelligent Vehicles (PPNIV) at the IEEE Int. Conf. Intell. Robots Syst. (IROS)}, 2020.
}
}

\begin{document}
\copyrightstatement
\maketitle
\thispagestyle{empty}
\pagestyle{empty}

\maketitle

\begin{abstract}

Decision making under uncertainty can be framed as a partially observable Markov decision process (\pomdp).
Finding exact solutions of \pomdps is generally computationally intractable, but the solution can be approximated by sampling-based approaches.
These sampling-based \pomdp solvers rely on multi-armed bandit ({\mab}) heuristics, which assume the outcomes of different actions to be uncorrelated.
In some applications, like motion planning in continuous spaces, similar actions yield similar outcomes.
In this paper, we utilize variants of \mab heuristics that make Lipschitz continuity assumptions on the outcomes of actions to improve the efficiency of sampling-based planning approaches.
We demonstrate the effectiveness of this approach in the context of motion planning for automated driving.
\end{abstract}

\IEEEpeerreviewmaketitle
\section{Introduction}
\label{introduction}

Sequential decision making problems in which the system dynamics are uncertain and the system state is unobservable can be framed as a \pomdp.
The \pomdp framework encodes uncertain and incomplete knowledge not by single states, but by beliefs over all possible states.
By optimizing over a sequence of actions and observations, it considers a very large number of possible future outcomes.
This comes at the cost of high computational complexity.

A \pomdp can typically be solved either by performing \emph{value iteration}, or by \emph{Monte Carlo tree search}.
The latter are real-time capable and are more flexible since they do not require state discretization. 
Common approaches are the algorithms \pomcp \cite{silver2010monte} and \tapir \cite{klimenko2014tapir}.
They employ the Upper Confidence Bound ({\ucb}) \cite{auerFinitetimeAnalysisMultiarmed2002} bandit algorithm to explore promising actions and exploit good ones.

\ucb does not make any assumption on the outcomes of similar actions.
This is essential for certain decision making problems, e.g.\ playing board games.
However, many real-world applications operate on compact, continuous spaces in which the profile of outcomes are continuously differentiable.
Furthermore, the uncertainties present in the environment have a \textit{smoothing effect} on any discontinuity, if they can be represented with a probability distribution whose density is continuous.
Such applications can benefit from Lipschitz continuity assumptions on the outcomes of actions.

A {\pomdp} solver can exploit the dependence between the expected rewards of different actions by utilizing a \mab that assumes Lipschitz continuity.
Even though this would substantially improve the efficiency of the sampling, currently there is no {\pomdp} solver that exploits this property.
In this paper, we propose to utilize a Lipschitz \mab that can work within the POMDP framework.
We further investigate whether the motion planning problem in automated driving has a continuous reward profile by analyzing the dependencies between actions in different scenarios, including edge cases like collisions.

In order to highlight the contribution of this work, we first provide background information on {\pomdps} in Section~\ref{sec:background}.
Next, in Section~\ref{sec:driving_pomdp} we focus on existing works that deal with motion planning for automated vehicles by utilizing the {\pomdp} framework and subsequently introduce our model, which shows several differences to existing works.
Once the underlying settings are introduced, we provide an overview on {\mab}s we benchmark in Section~\ref{multi-armed_bandits}.
We use a modified version of the POMCP algorithm to efficiently solve the POMDP problem,
presenting those modifications in Section~\ref{sec:pomcp_mod}, before analyzing the structure of reward profiles and benchmarking the {\mab}s in the Evaluation section.

\section{Background}
\label{sec:background}

A \pomdp operates on spaces of states $\stateSpace$, observations $\observationSpace$ and actions $\actionSpace$.
The state $\state$ cannot be observed exactly, and therefore, it is represented as a probability distribution over the state space, i.e.\ belief $\belief{(\state)}$.
Transitions from one state to another and their respective observations are given by a transition and an observation model.
Every time the agent chooses an action $\action \in \actionSpace$ and the environment transitions to the next state, 
the agent receives the scalar reward $r \in  \rewardSpace$, computed by the reward function $\rewardModel (s, a)$.
The action-value $\qvaluePolicy{} (\belief, \action)$ represents the expected future reward when taking action $\action$ in belief $\belief$ and following policy $\policy$ afterwards.

The \pomcp algorithm \cite{silver2010monte} applies Upper Confidence Bound for Trees (UCT) \cite{kocsisBanditBasedMonteCarlo2006} to \pomdps by iteratively sampling sequences of states and observations.
These sequences are kept track of in a tree data structure of nodes and edges, corresponding to states, actions and observations.
Key to the \pomcp algorithm is the equivalence of belief $\belief$ and history $\history$. 
A history is the sequence of actions and observations that have been selected and observed by the agent,
${\history}_t = \{{\action}_1 , {\observation}_1, \ldots, {\action}_t , {\observation}_t \}$.
When the initial belief ${\belief}_0$ is fixed and known, the belief is represented by the history of the tree.

One episode of the sampling procedure samples a single particle and involves four phases:
In the \texttt{simulation} phase of {\pomcp}, actions are selected by the \mab algorithm. 
Based on the generative observation model, the next state is determined.
When simulation reaches a node that is not in the tree yet, the tree is \texttt{expanded} by the node and the simulation stops.
The initial value estimate of the new node is done by \texttt{rollout} policy, which is used as an heuristic in lieu of expanding the tree further.
In the final phase, \texttt{backup}, the encountered nodes are updated from bottom-up in light of the newly gathered information.

\section{Motion Planning in Automated Driving with the \pomdp Framework}
\label{sec:driving_pomdp}

In automated driving, motion planning has to consider uncertain information such as the unknown intentions and future directions of other traffic participants, or objects in occluded areas.
One way to consider these uncertainties is to use the \pomdp framework.

\subsection{Related Work}
\pomdps have been used in previous works to solve the motion planning problem for automated driving.
Bai \etal \cite{baiOnlineApproachIntersection2014} use the \pomcp algorithm for planning in an intersection scenario.
They model the uncertain intentions of other drivers and driving style.
Although the \pomcp solver is capable of handling continuous state spaces, the authors used discrete states.
Brechtel introduces a \pomdp solver that is able to work with continuous spaces by learning a problem-specific representation of the state space \cite{brechtelProbabilisticDecisionmakingUncertainty2014}, \cite{brechtel2015dynamic}.

The importance of considering uncertainties in the planning of motion for automated driving is shown by Sunberg \etal \cite{sunberg2017value}.
They compare different planning frameworks based on MDPs and \pomdps, which are evaluated on a lane change scenario.
The results clearly indicate the need to take uncertainties into consideration.
For the \pomdp variant the \pomcp algorithm is enhanced with a technique called double progressive widening, enabling its use in continuous spaces \cite{bouton2017belief, sunberg2018online}.

Sefati \etal \cite{sefati2017towards} also include uncertainties in the motion planning for an intersection scenario.
They consider the unknown intentions of other traffic participants by inferring the state with a Bayesian network model. 
To solve the \pomdp, the MCVI solver \cite{bai2010monte} is combined with ideas form the SARSOP algorithm \cite{kurniawati2008sarsop}.
Their approach uses several heuristics to guide the exploration through the action space, though sparse, containing only three actions.

Bouton \etal \cite{boutonScalableDecisionMaking2018} study intersection and pedestrian crosswalk scenarios.
They do not deal with intentions of other road users, but focus on the integration of occlusions.
The action space includes only four or five actions, depending on the scenario.
The \pomdp problem with multiple participants is split into many problems involving only one other agent.

Hubmann \etal \cite{hubmannAutomatedDrivingUncertain2018} use a real-time method for computing solutions to intersection scenarios with multiple vehicles with unclear intentions.
They use the \tapir algorithm and consider an action space of only four actions.
In a later work they extend their approach to tackle occlusions \cite{hubmann2019pomdp}.
\subsection{Modeling the Motion Planning Problem for the \pomdp Framework}
\label{motion_planning}

The POMDP framework requires model-based representations on which to operate.
We closely follow the model presented in \cite{hubmannAutomatedDrivingUncertain2018} with some modifications to work better with a denser action space.

\subsubsection{Map data}

We refer to a path and its accompanying data as a \emph{route} and denote it by $\route$.
Accompanying data consists of the curvature $\curvature$ and the reference velocity $\speed$. 
In this way, we store every route $\rho$ as a tuple of $n$ points
\begin{equation}\nonumber
\rho = ({\routeData}_i)_{i = 1, \dots, n} \quad \text{with} \quad {\routeData}_i = ({\positionCartesianX}_i, {\positionCartesianY}_i, {\positionFrenetLon}_i, {\curvature}_i, {\speed}_{i}) ^ \top \in \mathbb{R}^5,
\end{equation}
where ${\positionCartesianX}$ and ${\positionCartesianY}$ are Cartesian coordinates and ${\positionFrenetLon}$ is the distance along the path.

\subsubsection{States, Observations, Actions}
We apply path-velocity decomposition \cite{kantEfficientTrajectoryPlanning2016} to reduce the complexity and, therefore, we plan acceleration along a path.
The state $\state$ of a vehicle $k$ is given by $\state_k = ({\positionFrenetLon}_k, {\speed}_k, {\rho}_k)$. %
For the ego vehicle the route $\route$ is known, hence, we represent its state as ${\state}_0 = ({\positionFrenetLon}_0, {\speed}_0)$.
For $k$ other vehicles in a traffic scene, we have the combined state $\state = ({\state}_0, {\state}_1, {\state}_2, \dots, {\state}_k)$. 

Observations ${\observation}$ are defined in Cartesian coordinates and only contain information about $k$ other traffic participants ${\observation} = ({\observation}_1, {\observation}_2, \dots, {\observation}_k)$ with ${\observation}_k = ({\positionCartesianX}_k, {\positionCartesianY}_k, {\speed}_k)^\top \in \mathbb{R}^3$. 

Actions available to the planner represent discrete values of acceleration $\action$.

\subsubsection{Transition model}
The \pomdp model is discretized in time and the transitions have the timestep $\samplingInterval = 1.0 \si{\second}$.
The vehicles follow a constant acceleration model and the route does not change.
For the ego vehicle the route is predefined and the acceleration input ${\action}_0$ is simply the action chosen by the \pomdp solver.

For predicting accelerations of other vehicles ${\acceleration}_k$, we use the Intelligent Driver Model (IDM) \cite{treiberCongestedTrafficStates2000}.
This model consists of a free driving term ${\acceleration}_{\text{ref},k}$ that allows the vehicle to smoothly adjust the velocity in case of deviations from its reference, and an interaction term ${\acceleration}_{\text{int},k}$ for avoiding collisions.
We saturate the resulting acceleration value with the maximum deceleration $\decelerationMax$ and add Gaussian noise $a_{\text{noise},k} \sim \mathcal{N}(0, \sigma^{2}_{a})$ to cover for modeling errors and uncertainties in the behavior model. %
\begin{equation}\nonumber
{\acceleration}_k = \max ( {\acceleration}_{\text{ref},k} + {\acceleration}_{\text{int},k}, \ \decelerationMax ) + {\acceleration}_{\text{noise},k}.
\end{equation}

IDM may yield accelerations resulting in negative velocity due to time discretization.
In such cases, we calculate the stop position from the motion kinematics.

\subsubsection{Observation model}
The observation model generates a possible observation from a given state-particle.
After the transformation from path coordinates to Cartesian coordinates, we add observation noise sampled from independent Gaussian distributions.

\subsubsection{Reward model}
\label{subsec:reward_model}
The terms in the reward model resemble those of an ordinary trajectory optimization problem.
All terms are modeled as negative rewards
\begin{equation}\nonumber
\reward = {\reward}_{\text{coll}} + {\reward}_{v} + {\reward}_{j,\text{lon}} + {\reward}_{a, \text{lat}}.
\end{equation}
The collision term ${\reward}_{\text{coll}}$ takes only collisions of the ego vehicle into account
\begin{equation}\nonumber
{\reward}_{\text{coll}} =
		\begin{dcases*}
        0  & no collision \\
        \costScalingFactor_{\text{coll}} & ego vehicle collides
        \end{dcases*}.
\end{equation}
The criterion from \cite[p.\ 223]{ericson2004real} is applied for the collision check, which is performed during state transition.
The corresponding reward therefore is a function of the current state and the posterior one, \ie $\rewardModel (s, a, s^{\prime})$.

We punish the deviation of the ego vehicle velocity from a predefined reference with an asymmetric loss function
\begin{equation}\nonumber
{\reward}_{v} =
		\begin{dcases}
        \costScalingFactor_{{\speed}} \ ({\speed}_0 - \speedReference)^2 & \text{if} \ {\speed}_0 \geq  \speedReference \\
        \costScalingFactor_{{\speed}} \ \log \! \left( 1 + \left( {\speed}_0 - {\speedReference} \right)^2 \right) & \text{otherwise}
        \end{dcases},
\end{equation}
with $\costScalingFactor_{\speed}$ being the cost scaling factor. 
We apply quadratic cost if $\speedReference$ is exceeded and \emph{Cauchy loss} \cite{barronGeneralAdaptiveRobust2017} otherwise.  
The asymmetric loss function is motivated by the assumption, that driving slow or standing still might be part of an ordinary solution to the motion planning problem, whereas driving with higher speed is only acceptable in extraordinary cases. %

The third term in the reward model ${\reward}_{{\jerk},\text{lon}} = \costScalingFactor_{{\jerk},\text{lon}} {j_0}^2$ accounts for jerk and considers changes in the longitudinal acceleration.
The last term is the lateral acceleration term ${\reward}_{\acceleration,\text{lat}} = \costScalingFactor_{\acceleration,\text{lat}} \left( \curvature \ {\speed}^{2}_{0} \right)^2$.

\vspace{-0.5mm}\section{Multi-armed bandits}
\label{multi-armed_bandits}

The general \mab is a sequential decision problem \cite{slivkins2019introduction}, which proceeds in rounds denoted by $t \in \lbrace 1, \dots, \banditRounds \rbrace$.
At every round, the agent picks one arm $ \banditArm $ from some set of arms $ \banditArmSet = \lbrace \action_1, \action_2, \ldots, \action_K \rbrace $, where $K$ denotes the number of arms,
with the goal of maximizing the rewards it collects over $ T $ rounds.
In the stochastic setting, every arm corresponds to a reward distribution, whose properties are unknown to the agent.
The distributions are assumed to be stationary, \ie they do not change over time.
The \mab algorithm balances between the exploration of unknown arms and exploitation of good arms.
Several approaches have been suggested in the past to deal with the \mab problem.
\subsection{UCB}
Introduced by Auer \etal \cite{auerFinitetimeAnalysisMultiarmed2002}, UCB is \textit{optimistic in the face of uncertainty}.
Its robustness and practicality have been proven in numerous applications.
In each round $t$ the algorithm calculates the index
\begin{equation}\label{eq:ucb}
{\belief}_t(\banditArm) = \hat{\mu}_t(\banditArm) + \explorationConstant \, \sqrt{\frac{2 \log t}{n_t(\banditArm)}}
\end{equation}
for every arm and then chooses the arm with the highest index (\cf Algorithm~\ref{alg:ucb}).
The first term is the current average reward of an arm $ \banditArmMeanReward{\estimator}{t} $.
The second term is called the confidence radius and is proportional to the upper bound of the confidence interval of the average reward. %
The denominator $ n_t(\banditArm) $ is the number of times arm $ \banditArm$ has been played and $ \explorationConstant \in \realNumbers$ is the \emph{exploration constant} trading-off exploitation and exploration.

\vspace{4mm}
\begin{algorithm}
    \DontPrintSemicolon
    \BlankLine
    \uIf{$ t \leq K $}{
        Choose arm from $ \lbrace \banditArm:n_t(\banditArm) = 0 \rbrace $ at random
    }
    \Else{
        Choose arm $a_t = \argmaximum{\banditArm \in \banditArmSet} b_t(a)$
    }
    \BlankLine
    \caption{Upper Confidence Bound (UCB)}
    \label{alg:ucb}
\end{algorithm}

A more general case of {UCB} is the {KL-UCB} bandit algorithm.
It allows distributions of canonical exponential family to be set for the reward.
If rewards of the arms are assumed to be Gaussian distributed with equal variance, 
the {KL-UCB} index becomes equal to Equation~\ref{eq:ucb}, but with an exploration constant of $ {\explorationConstant = \sqrt{\sigma^2}} $.

\subsection{UCB-V}
Audibert \etal enhanced \ucb by including the estimated reward variances $ \hat{\sigma}_t^2(a) $ \cite{audibertTuningBanditAlgorithms2007}.
In contrast to {\ucb}, the variances of the rewards are not assumed to be equal.
The index is given as
\begin{equation}\label{eq:ucb-v}
b_t(a) = \hat{\mu}_t(a) + \sqrt{\frac{2 \hat{\sigma}_t^2(a) \log t}{n_t(a)}} + \frac{3 \explorationConstant \log t}{n_t(a)}.
\end{equation}
\ucbv still has the exploration constant in the third term, but with growing $ n_t(a) $ the exploration constant loses influence and the estimated variances are more strongly considered.
To be precise, the second term has $ \sqrt{n_t(a)} $ times the weight of the third term.

\subsection{POSLB}

The Pareto Optimal Sampling for Lipschitz Bandits (\poslb) algorithm assumes the expected rewards to be Lipschitz continuous and thereby improves the efficiency of sampling by guiding the bandit faster to more rewarding arms \cite[p.\ 21]{magureanu2018efficient}.
Although Lipschitz continuity in the reward function values is assumed, the set of arms is still discrete.
The expected reward function over arms is assumed to obey
\begin{equation}\nonumber
\left| \mu(a) - \mu(a^{\prime}) \right| \leq \lipschitzConstant \left| \banditArm - \banditArm^{\prime} \right|
\end{equation}
for any pair of arms $ (\banditArm, \banditArm^{\prime}) $ and a Lipschitz constant $ \lipschitzConstant $.
The {\poslb} algorithm for Kullback-Leibler divergence of Gaussian distributed rewards is given in Algorithm~\ref{alg:poslb}.
The algorithm identifies the currently best arm $a^{\ast}_t$ and calculates the KL-UCB index $b_t(a^{\ast}_t)$.
The intermediate values $ \bigl( \lambda_t(a, a_1), \dots, \lambda_t(a, a_K) \bigr) $ can be understood as the \emph{most confusing} estimated reward vector, which would make the suboptimal arm $a$ the optimal one.
The Lipschitz assumption is integrated into the bandit via the confusing rewards and is adjusted by $\lipschitzConstant$.
\poslb looks at the differences between the most confusing rewards and the actual estimated rewards and favors arms where the sum of the differences, weighted by their visit counts, is small.

\vspace{4mm}
\begin{algorithm}
	\DontPrintSemicolon
	\BlankLine
	\uIf{$ t \leq K $}{
		Choose arm from $ \lbrace \banditArm:n_t(\banditArm) = 0 \rbrace $ at random
	}
	\Else{
        $ \optimum{\banditArm}_t = \argmaximum{\banditArm \in \banditArmSet} \banditArmMeanReward{\estimator}{t} $\\  %
		$ f_t(a) = \vspace{2mm} \hspace{-3mm}
		\begin{dcases}
		\displaystyle\sum_{\banditArm^{\prime} \in \banditArmSet} n_t(\banditArm^{\prime}) \bigl( \banditArmMeanReward[{\banditArm}^{\prime}]{\estimator}{t} - \lambda_t(\banditArm, \banditArm^{\prime}) \bigr)^2 (2 \sigma^2)^{-1} & \hspace{-1mm} \text{if} \ \banditArm \neq \optimum{\banditArm}_t \\
    	n_t(\banditArm) \bigl( \banditArmMeanReward{\estimator}{t} - b_t(\optimum{\banditArm}_t) \bigr)^2 (2 \sigma^2)^{-1} & \hspace{-1mm}  \text{if} \ \banditArm = \optimum{\banditArm}_t
		\end{dcases}
		$\\[2mm]
		with $ \lambda_t(\banditArm, \banditArm^{\prime}) = \max \bigl( b_t(\optimum{\banditArm}_t) - \lipschitzConstant \left| \banditArm - \banditArm^{\prime} \right|, \ \banditArmMeanReward[{\banditArm}^{\prime}]{\estimator}{t} \bigr) $\\[2mm]
		Choose arm $\banditArm_t = \argmaximum{\banditArm \in \banditArmSet}{ \log t - f_t( \banditArm ) }$
	}
	\BlankLine
	\caption{POSLB}
	\label{alg:poslb}
\end{algorithm}

The algorithm runs with an increased complexity of $\complexity (\abs{\banditArmSet}^2) $ compared to \ucb.
Note that, if the rewards are Gaussian distributed, the variance of POSLB is equal to the selected $\explorationConstant$ of UCB.
Hence, $\explorationConstant$ has a similar impact on both.

\subsection{POSLB-V}
While \poslb is able to consider the Lipschitz continuity of the reward function, it does not make use of estimated variances.
Simply replacing the variance parameter $\sigma^2$ in the \poslb algorithm by the estimated variances ${\estimator{\sigma}}^2(\banditArm)$ leads to poor results.
Instead, it is advisable to shift from exploration constant to estimated variances over time, like it is done in \ucbv.

Equating the {KL-UCB} and \ucbv indices and solving for the variance parameter leads to
\begin{equation}\nonumber
\sigma^2_t(\banditArm) = \sigma^2 = \frac{n_t(\banditArm)}{2 \log t} \left( \sqrt{\frac{2 \hat{\sigma}_t^2(\banditArm) \log t}{n_t(\banditArm)}} + \frac{3 \explorationConstant \log t}{n_t(\banditArm)} \right)^2.
\end{equation}
We then use the estimated variances and Lipschitz assumption together and and call it \poslbv.

The augmented variance $\sigma^2_t(a)$ is then used in \poslb to calculate $b_t(a^{\ast}_t)$ and $f_t(a)$, instead of the fixed $\sigma^2$.
Otherwise, the algorithm stays unchanged.

\subsection{Continuous bandits}
All bandits above need to discretize the continuous action space.
However, there are several bandit algorithms that can handle continuous action spaces \cite{mansley2011sample,weinstein2012bandit,bull2015adaptive}.
Other {\mab}s additionally assume that the returns of the arms are Lipschitz continuous \cite{kleinberg2008multi,bubeck2009online,maillard2010online,bubeck2011lipschitz}.
An overview is provided in \cite[p.\,40]{slivkins2019introduction}.
In every round they choose a new arm from the action space, never sampling the same action twice.
This prohibits the usage of these bandits within the \pomcp algorithm, which needs to build a belief tree.
The bandit in \cite{wang2017cemab} can deal with a large amount of discrete actions, but cannot easily be utilized in a belief tree.
None of these approaches is compatible with real-time \pomcp and, therefore, are not investigated further.

\section{Solver configurations}
\label{sec:pomcp_mod}

We adapt the \pomcp algorithm \cite{silver2010monte} by modifying the \texttt{simulation}, \texttt{rollout}, and \texttt{backup} phases.

In the \texttt{simulation} phase of {\pomcp}, we compare the {\mab}s presented in the previous sections for choosing actions. 
In the \texttt{rollout} phase we perform constant velocity rollouts for their minimal computational time.

We modify the \texttt{backup} phase in several ways. 
We use incremental statistics \cite{finch2009incremental} to calculate the mean and the variance of Q-values
\begin{align}
\qvalue_n &= \qvalue_{n-1} + \learningRate(n) \ \bigl( \widehat{\qvalue} - \qvalue_{n-1} \bigr), \\
\sigma^2_n &= \bigl( 1 -  \learningRate(n) \ \bigr) \Bigl( \sigma^2_{n-1} + \operatorname{\learningRate}(n) \bigl(\widehat{\qvalue} - \qvalue_{n-1} \bigr)^2 \Bigr).
\end{align}
We alter the learning rate $\learningRate(n) = \frac{1}{n^\omega} $ by choosing a \textit{polynomial learning rate} $\omega < 1$ instead of a \textit{linear} rate where $\omega = 1$.
As pointed out in \cite{even-darLearningRatesQlearning2003}, setting $\omega=0.77$ has theoretically superior convergence properties, which we also observed in our initial empirical analysis.
We use the maximum of Q-values as an estimate of the belief nodes value
\begin{equation*}
\Value (\history) = \maximum{\banditArm \in \banditArmSet} \ \qvalue( \history, \action).
\end{equation*}

The belief tree of the \pomdp solver works only with discrete observations.
Therefore, we discretize the continuous observation space in a data stream clustering manner:
A list of cluster centers is maintained and every new observation is compared to the entries in this list. 
If the Euclidean distance between observation and a cluster center is within a given threshold, the observation is assigned to the first matching cluster encountered in the ordered list. 
Otherwise, a new cluster center is inserted at the end of the list.
\section{Evaluation}
\label{results}

In our experiments, we take accelerations from the comfortable range $\action \in [\acc{-3}, \acc{1}]$ with equidistant spacing as available actions to the planner.
Hence, discretization of $5$, $9$, $17$, and $33$ actions correspond to acceleration intervals $ \Delta \action$ of $\acc{1}$, $\acc{0.5}$, $\acc{0.25}$, and $\acc{0.125}$ respectively.
Parameter values used in evaluation are provided in \textsc{Appendix}.

\begin{figure}[!h]
\centering
\begin{subfigure}{.5\columnwidth}
  \centering
  \includegraphics[width=.485\columnwidth]{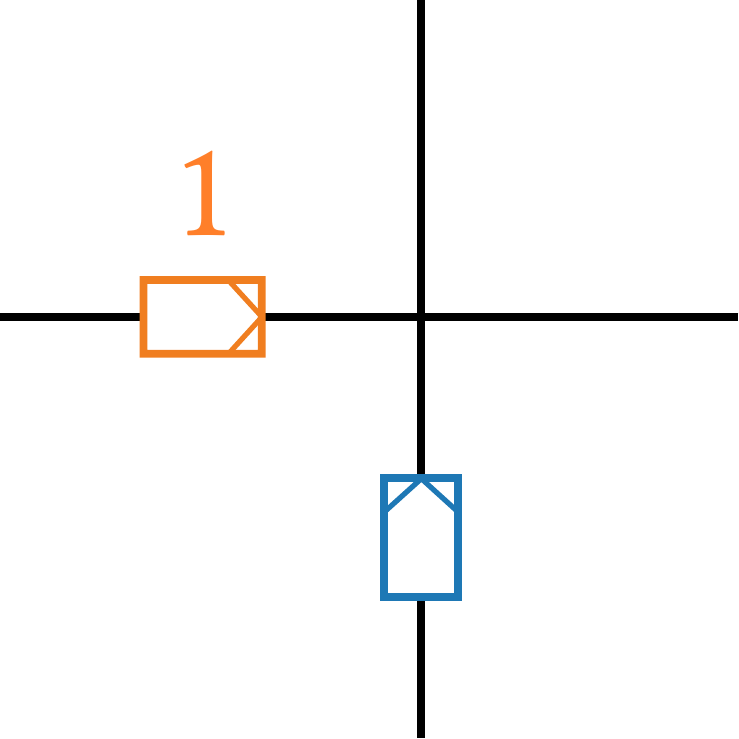}
  \caption{Collision scene.}
  \label{fig:results-scenario-description-c}
\end{subfigure}%
\begin{subfigure}{.5\columnwidth}
  \centering
  \includegraphics[width=.485\columnwidth]{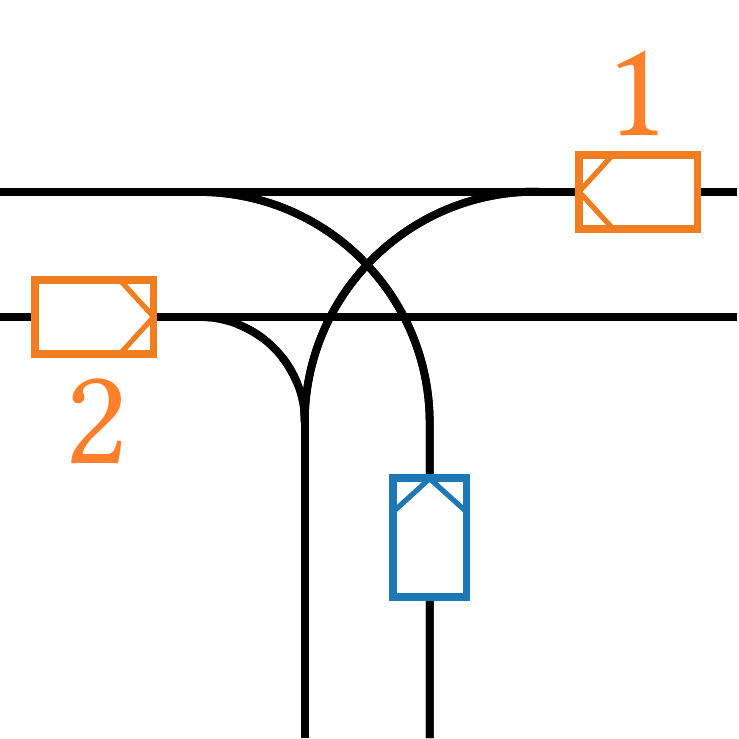}
  \caption{Intersection scene.}
  \label{fig:results-scenario-description-i}
\end{subfigure}
\caption{Traffic scenes used in the evaluations.}
\label{fig:results-scenario-description}
\end{figure}

We used two simple traffic scenarios "straight driving" and "traversing curves" for initial testing.
For evaluation, we use two complex traffic scenes in which interaction with other participants are required (cf. Fig.~\ref{fig:results-scenario-description}).
As the solution depends on the initial belief, we sample the state-particles of the initial belief from the same probability distribution.
We assume that the distributions of position, velocity and route are independent.

Collisions pose a counter-example to the underlying assumption in this work: they introduce an abrupt change in the reward and hence pose the most challenging problem.
In the collision scenario \collisiontwo{} we simulate the case of an imminent collision.
In the intersection scene, the ego vehicle has to identify whether the other vehicles are on collision paths and avoid collisions.
We define two scenarios based on this scene: the \mczero{} scenario and the \mctwo{} scenario, which pose low and high probabilities of collision, respectively.
The parameters of the scenarios are provided in \textsc{Appendix}.

\subsection{Q-value Function Analysis}

If the transition model and reward function of the ego vehicle are Lipschitz continuous, the resulting Q-value function of an MDP is guaranteed to be Lipschitz continuous as well \cite{asadi2018lipschitz}.
To expand the result to {\pomdp}s, intuitively the observation function needs to be Lipschitz continuous as well.
This is not the case in our model, as the reward includes the binary collision term, rendering the Q-value function discontinuous.
We argue, however, that the noise, which is present in the transition and observation model, has a smoothing effect on these discontinuities.

In order to empirically analyze the continuity of Q-value function, we accurately evaluate it in the root node of the belief tree.
We set up equidistant actions with $ \Delta \action = \acc{0.05} $ and thereby cover the action space densely.
Then, the Q-values of these actions are evaluated by conducting a simulation run for each of the actions with $10^6$ particles.
During a single run, the action in the root node is kept fixed, whereas in the following belief nodes {\ucb} is used to select among five available actions.

\begin{figure}[!ht]
\includegraphics{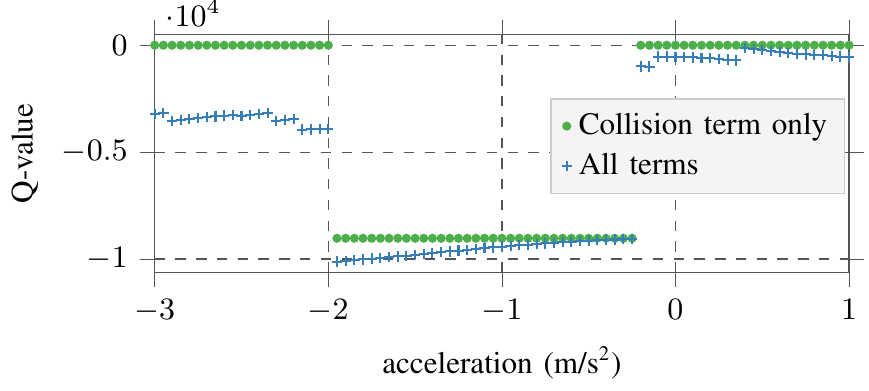}
\vspace{-\baselineskip}
\caption{Q-value profile of \collisiontwo{} without uncertainty. \vspace{-2mm}}
\label{fig:results-smooth-collision-collision2-no-noise}
\end{figure}

To highlight the discontinuity in the reward, we evaluate \collisiontwo{} without uncertainties (cf.\ Fig.~\ref{fig:results-smooth-collision-collision2-no-noise}).
If the agent chooses an action between \acc{-2.0} and \acc{-0.2} a collision cannot be prevented. 
The step-like patterns emerge due to discretized values of braking actions in subsequent timesteps.

 \vspace{0mm}

\begin{figure}[!ht]
\vspace{2mm}
\includegraphics{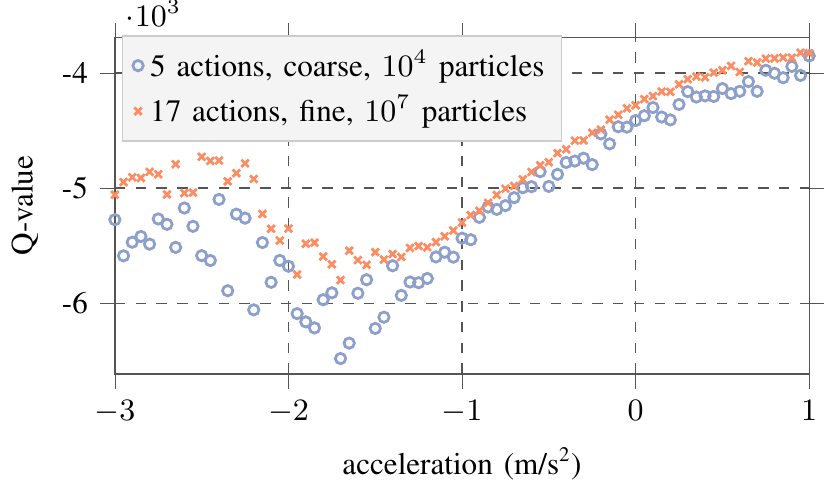}
\caption{Approximations of the Q-value profile of {\collisiontwo}. \vspace{-2mm}}
\label{fig:results-smooth-collision-collision2}
\end{figure}

Coarse and more accurate approximations of the Q-value profiles of \collisiontwo{} are presented in Fig.~\ref{fig:results-smooth-collision-collision2}.
The coarse one is approximated with \num{e4} particles, whereas the more accurate one is obtained by sampling \num{e7} particles with $17$ actions, and doubled resolution of the observation discretization.
The Q-value profile converges to a continuous function.
From the more accurate simulation, it can be seen that the variance is reduced.
However, as a result of reduced smoothing, the variance between successive actions tend to be higher than the rest at discontinuities.
This points out that the Lipschitz assumption loses its validity for tens of millions of samples under low uncertainty.

\begin{figure}
\vspace{2mm}
\includegraphics{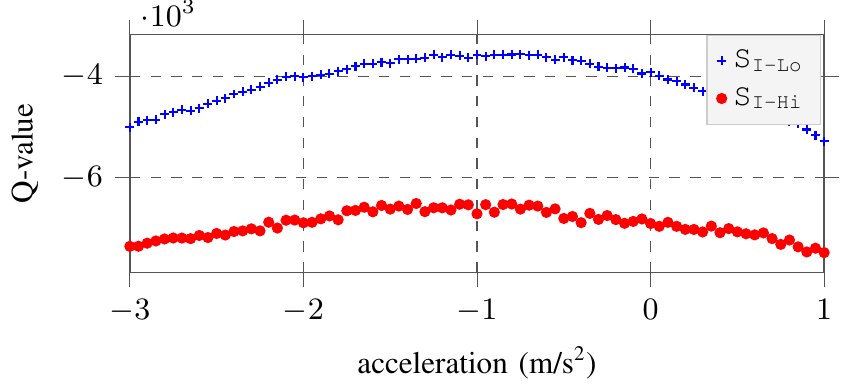}
\resizebox{\columnwidth}{!}{
}	
\vspace{-8mm}
\caption{Approximated Q-value profiles of the intersection scenarios sampled with $10^4$ particles. \vspace{-6mm}}
\label{fig:results-smooth-intersection-mc0}
\end{figure}

The Q-value profile of \mczero{} and \mctwo{} are presented in Fig.~\ref{fig:results-smooth-intersection-mc0}.
Both are very smooth and have the same underlying shape, whereas the overall level of Q-values is less and the variances are higher in {\mctwo}, as a result of the higher collision probability.

The continuity evaluation illustrates that the uncertainties partially smooth the discontinuities. %
Increasing the number of available actions increases the smoothness of the profile, as well. 
Notice that we model uncertainties in the longitudinal direction only. 
Considering the lateral position uncertainty would further smooth the Q-values.
\vspace{1mm}

\subsection{Convergence Analysis}

A standard metric to benchmark convergence is to identify the optimal action after a predefined number of samples $n$. 
We use the \textit{mean absolute error} (MAE) between the current best action $\optimum{\action}_n$ and the optimal action $\optimum{\action}$  as a performance measure.

We calculate $\optimum{\action}$ by sampling \num{e7} particles while employing {\ucb} bandit.
The observed Q-values still have stochastic nature and hence, we perform Gaussian process regression to eliminate the noise of the Q-values and to recover the underlying function. 
As the profile of the Q-values is sufficiently smooth,
we use a squared exponential kernel with length scale and noise level as hyperparameters.
We estimate the optimal action for every scenario and number of available actions by evaluating the mean of the fitted Gaussian process at the locations of the available actions.
The results are given in TABLE~\ref{tab:optimal-actions} in \textsc{Appendix}.

To perform a reliable analysis, we calculate the MAE over multiple simulation runs $m$
\begin{equation}\nonumber
\mathrm{MAE}_n = \frac{1}{m} \sum_{i = 0}^{m-1} \abs{\optimum{\action}_{i, n} - \optimum{\action}},
\end{equation}
where $m=100$ and ${2 \cdot 10^4}$ episodes. 
Given the ground truth, we determine $\optimum{\action}_n$ as 
\begin{equation}\nonumber
\optimum{\action}_n = \argmaximum{\banditArm \in \banditArmSet} \ \qvalue_n(\history_0, \action).
\end{equation}

The optimal Lipschitz constant required for \poslb is not known in advance.
Overestimating and underestimating lead to suboptimal performance.
We use the mean of the fitted Gaussian process and empirically select the Lipschitz constant $ \lipschitzConstant = \num{2000} $ for these scenarios (cf. TABLE~\ref{tab:lipschitz-constant} in \textsc{Appendix}).

\vspace{2mm}
\begin{figure}[t!]
	\centering
	\begin{subfigure}{\columnwidth}
		\centering
        \includegraphics{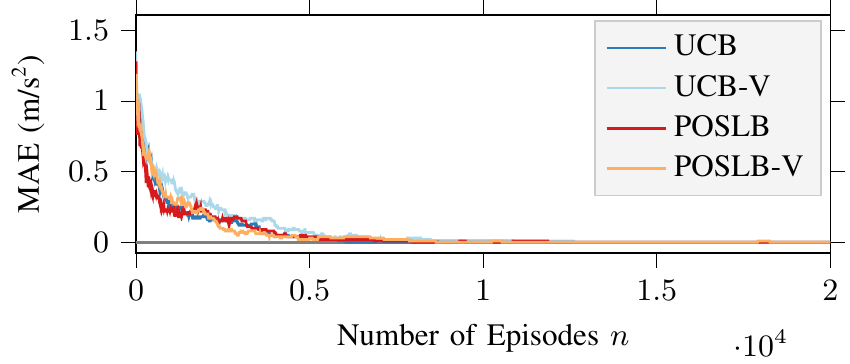}
        \vspace{-7mm}
		\caption{5 actions.}
        \vspace{1mm}
		\label{fig:results-log-mc2-action-dist-collision-a5}
	\end{subfigure}
	\begin{subfigure}{\columnwidth}
		\centering
        \vspace{2mm}
        \includegraphics{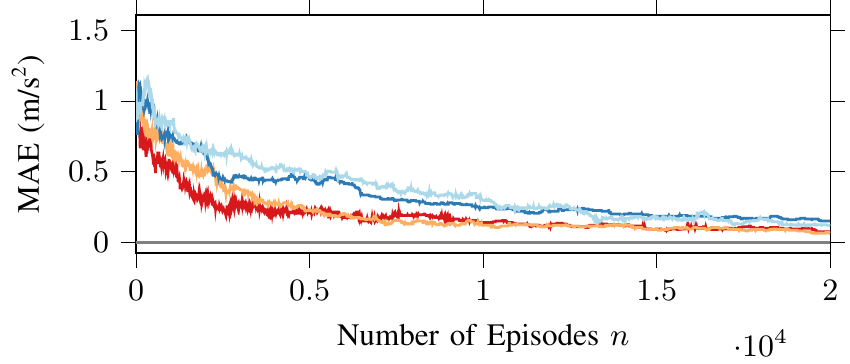}
        \vspace{-7mm}
		\caption{9 actions.}
        \vspace{1mm}
		\label{fig:results-log-mc2-action-dist-collision-a9}
	\end{subfigure}
	\begin{subfigure}{\columnwidth}
		\centering
        \vspace{2mm}
        \includegraphics{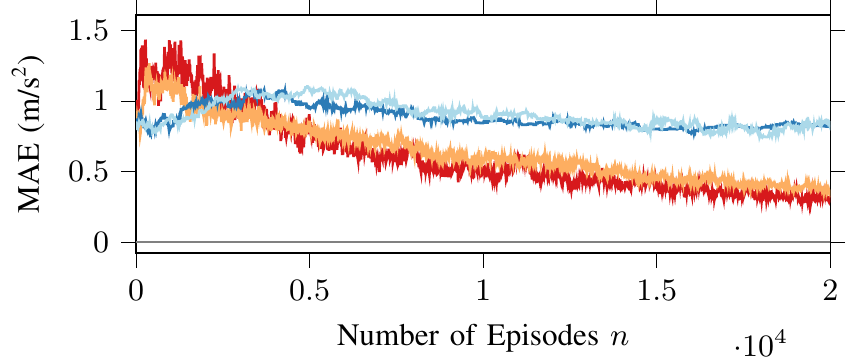}
        \vspace{-7mm}
		\caption{33 actions.}
		\label{fig:results-log-mc2-action-dist-collision-a33}
	\end{subfigure}
	\caption{\small Mean absolute error for {\collisiontwo}, for different numbers of actions. \vspace{-4mm}}
	\label{fig:results-log-mc2-action-dist}
\end{figure}
\begin{figure*}
	\begin{subfigure}[b]{0.45\textwidth}
		\centering 
        \includegraphics{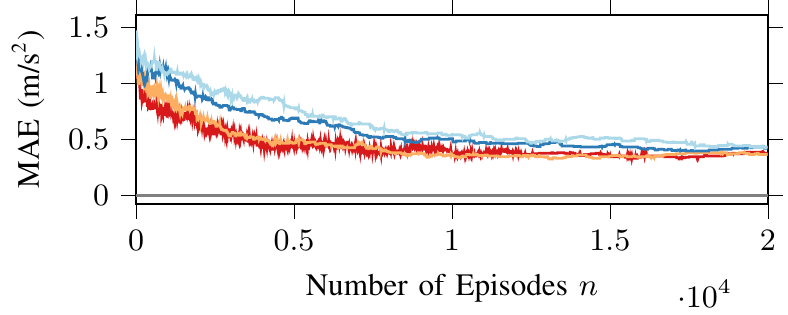}
        \vspace{-6mm}
		\caption{\small {\mctwo}, 9 actions.}
        \vspace{2mm}
		\label{fig:results-log-mc2-action-dist-a9}
	\end{subfigure} \hfil
	\begin{subfigure}[b]{0.45\textwidth}   
		\centering 
        \includegraphics{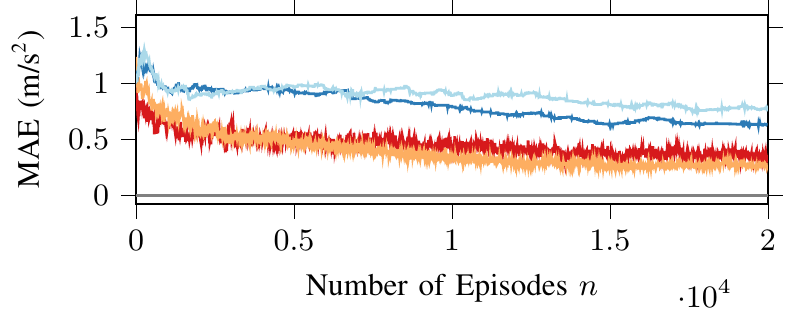}
        \vspace{-6mm}
		\caption[]%
		{{\small {\mctwo}, 33 actions.}}
        \vspace{2mm}
		\label{fig:results-log-mc2-action-dist-a33}
	\end{subfigure} \par\medskip
	\begin{subfigure}[b]{0.45\textwidth}  
		\centering 
        \includegraphics{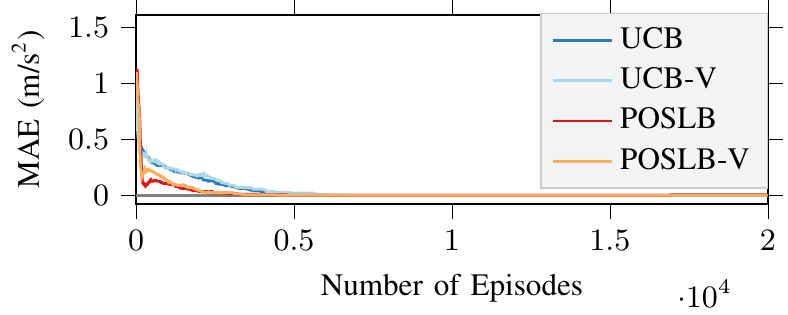}
        \vspace{-6mm}
		\caption[]%
		{{\small {\mczero}, 9 actions.}}    
        \vspace{2mm}
		\label{fig:results-log-mc0-action-dist-a9}
	\end{subfigure} \hfil
	\begin{subfigure}[b]{0.45\textwidth}   
		\centering 
        \includegraphics{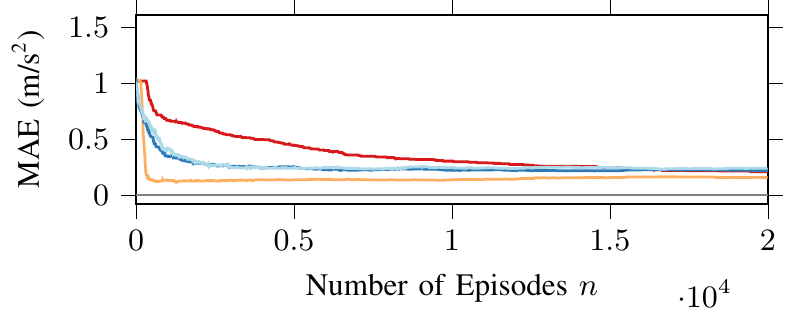}
        \vspace{-6mm}
		\caption[]%
		{{\small {\mczero}, 33 actions.}}    
        \vspace{2mm}
		\label{fig:results-log-mc0-action-dist-a33}
	\end{subfigure}
	\caption{Mean absolute error for the intersection scenarios for different number of actions.}
	\label{fig:results-log-mc0-action-dist}
\end{figure*}

Fig.~\ref{fig:results-log-mc2-action-dist} presents the convergence results for {\collisiontwo}.
From the figures, it is clear that for a low number of available actions all of the bandits have comparable convergence properties. 
However, as the number of actions increases, POSLB and POSLB-V show superior performance. 
In the case of $33$ actions, the UCB bandits have twice the MAE compared to Lipschitz bandits after $2 \cdot 10^4$ episodes. 
The variants considering variances perform comparably.
Another obvious result is that none of them can reach zero MAE.
The value they reach is equal to the action discretization $ \Delta \action$, as expected. 
Even though not presented, the results for $17$ actions lay between those for $9$ and $33$.
The results for $9$ and $33$ actions of \mczero{} are given in Figure~\ref{fig:results-log-mc0-action-dist}.
The results for \mctwo{} are very similar to those of \collisiontwo{} (cf.\ Fig.~\ref{fig:results-log-mc2-action-dist}), as expected.
Strikingly, the POSLB bandit shows the slowest convergence in the case of $33$ actions, and POSLB-V performs best. 
The poor performance of POSLB is caused by misleading rollouts which point to a different area of the action space to be optimal.
The Lipschitz assumption causes the bandit to select actions in that area.
By selecting actions with higher variances more often, POSLB-V compensates the drawbacks resulting from such misleading rollouts.
\mczero{} resembles such a narrow case in which the consideration of variances is advantageous.

\vspace{2mm}
 \vspace{0mm}

\subsection{Discussion}

The results of the convergence analysis present the average over $m=100$ runs.
We analyze the standard deviation of MAE (\stdevmae{}) for individual runs of different bandits.
The results indicate that the \stdevmae{} values are comparable,
whereas POSLB bandits have slightly smaller \stdevmae{} then their UCB counterparts.
The results for 9 actions in \collisiontwo{} are presented in TABLE~\ref{table:collision2-9actions-samples} in \textsc{Appendix} as an arbitrary example.

The noise-levels observed in the convergence plots are inversely proportional to the action discretization.
Optimal actions are typically selected more often when the discretization is bigger.
Therefore, convergence plots of increasing number of actions tend to be more noisy.

\begin{figure}
        \includegraphics{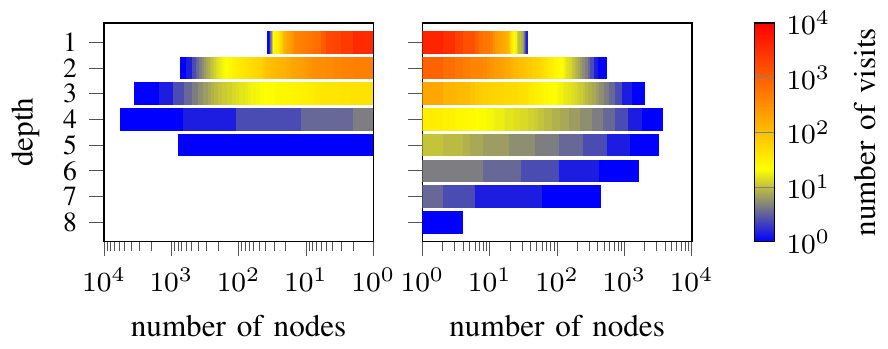}
    \vspace{-\baselineskip}
	\caption{Tree depth of UCB (left) and POSLB (right) bandits in {\collisiontwo} for the same number of episodes.}
	\label{fig:tree-depth}
\end{figure}

Tree depth of a solution is an important indicator of the quality: deeper trees consider longer horizons and are more accurate.
In Fig.~\ref{fig:tree-depth} we compare the tree depth for UCB and POSLB bandits for {\collisiontwo} with the same number of particles.
The bars in the figure represent the number of created nodes, and the color scales represent the number of visits for each level of the tree.
It is clear that UCB has a greater branching factor compared to POSLB, resulting in shallower trees.
In our experiments, other scenarios have verified this result.

Although the Lipschitz bandits have an increased computational complexity over UCB, we observed in our experiments that their runtime is never longer than $10\%$ longer.
This applies to the most demanding case of 33 actions.
On average, POSLB is $5\%$ slower.

\section{Conclusions}
\label{conclusion_future}

In this paper, we highlight the benefits of employing heuristics other than UCB in MCTS and present two important results.
As a first result, we empirically show that the uncertainty in the transition and observation models of the \pomdp formulation have a smoothing effect on the discontinuities in the Q-value function, eventually allowing for a Lipschitz continuity assumption.

We further show that the planning problem can be solved with fewer samples by utilizing the continuity of Q-values.
By replacing the standard multi-armed bandit (UCB) with one that assumes Lipschitz continuity (POSLB), considerable performance improvements are achieved for higher numbers of actions, especially in the early stages of sampling.

A further contribution is the POSLB-V bandit that is derived from the POSLB bandit.
Motivated from UCB-V, it considers the variance of Q-values during the action selection.
Experiments have shown that considering variances can be advantageous, in cases where the rollout policy might be misleading.

Real-time capability constraints have limited the use of existing \pomdps to decision making problems with fewer actions.
With this work, we are able to accelerate the speed of \pomdps with a novel tree expansion technique that exploits the Q-value structure of our problem.
This enables the use of \pomdps for problems where multiple actions need to be considered, such as in motion planning.

\section*{Appendix}

\begin{table}[H]
    \caption{Optimal action $ \optimum{\action} $ (\si{\metre\per\second\squared}).}
    \label{tab:optimal-actions}
    \centering

    \begingroup	
    \small
    \sisetup{table-number-alignment = center,
        table-figures-integer = 1,
        table-figures-decimal = 2}
    
    \begin{tabular}{S[table-figures-integer=2, table-figures-decimal=0]SSSSSSS}
        \toprule
        {$\abs{\actionSet}$} & {Straight} & {Curve} & {\collisiontwo{}} & {\mczero{}} & {\mctwo{}} \\
        \midrule
        5 & 0.0 & -1.0 & 1.0 & -1.0 & -1.0 \\
        9 & 0.0 & -1.5 & 1.0 & -1.0 & -1.5 \\
        17 & 0.0 & -1.5 & 1.0 & -1.0 & -1.25 \\
        33 & 0.0 & -1.0 & 0.875 & -0.875 & -1.125 \\
        \bottomrule
    \end{tabular}
    \endgroup
\end{table}

\begin{table}[H]
   	\caption{Estimated Lipschitz constant $ \lipschitzConstant $ (\si{\second\squared\per\metre}).}
	\label{tab:lipschitz-constant}
	\centering
	\begingroup	
	\sisetup{table-number-alignment = center,
		table-figures-integer = 3,
		table-figures-decimal = 0}
	
	\small
	\begin{tabular}{S[table-figures-integer=2, table-figures-decimal=0]SSSSSSS}
		\toprule
		{$\abs{\actionSet}$} & {Straight} & {Curve} & {\collisiontwo{}} & {\mczero{}} & {\mctwo{}} \\
		\midrule
		5 & 1247 & 1847 & 1003 & 1241 & 573 \\
		9 & 1271 & 2157 & 1981 & 1345 & 728 \\
		17 & 1336 & 2280 & 2742 & 1453 & 787 \\
		33 & 1370 & 2246 & 1260 & 1432 & 1033 \\
		\bottomrule
	\end{tabular}
	\endgroup
\end{table}

\begin{table}[h!]
    \small
    \caption{\stdevmae{} for 9 actions of {\collisiontwo}.}
    \label{table:collision2-9actions-samples}
    \begin{tabular}{llllllll} \toprule
        \multirow{2}{*}{Bandit}  & \multicolumn{7}{c}{Number of Episodes} \\ \cmidrule{2-8}
        & $10^0$   & $10^1$    & $10^2$   & $10^3$   & $10^4$   & $2 \cdot 10^4$   & $10^5$\\ \midrule
        UCB     & 0.57     & 0.57      & 0.81     & 0.55     & 0.29     & 0.23       & 0.07  \\
        UCB-V   & 0.52     & 0.52      & 0.66     & 0.53     & 0.32     & 0.21       & 0.09  \\
        POSLB   & 0.49     & 0.49      & 0.90     & 0.55     & 0.25     & 0.18       & 0.00  \\
        POSLB-V & 0.65     & 0.65      & 0.77     & 0.57     & 0.21     & 0.17       & 0.07  \\ \bottomrule %
    \end{tabular}
\end{table}

\begin{table}
	\caption{Times until the point-of-conflicts with different routes for the vehicles presented in the scenarios.}
	\label{table:results}
	\small
	\setlength\tabcolsep{0pt}
	\begin{tabular*}{\linewidth}{@{\extracolsep{\fill}} c l l c c c c c @{}}
		\toprule
		Scenario & & Vehicle & & \multicolumn{4}{c}{Time-to-Intersection (\si{\second})} \\
		\midrule
		\multirow{2}{*}{\collisiontwo{}} & & ego       & & 2.11 & & &  \\
		& & vehicle2  & & 2.71 & & &  \\
		\cmidrule{3-8}
		& & ego       & & 5.33 & 5.14 & 6.81 &      \\
		\mczero{} & & vehicle1  & & 3.99 & 4.20 & 4.78 &      \\
		& & vehicle2  & & 6.35 & 6.14 & 7.89 & 7.73 \\
		\cmidrule{3-8}
		& & ego       & & 2.66 & 2.28 & 5.63 &      \\
		\mctwo{}  & & vehicle1  & & 2.78 & 3.23 & 4.52 &      \\
		& & vehicle2  & & 3.42 & 3.05 & 6.21 & 5.92 \\                       
		\bottomrule
	\end{tabular*}
\end{table}

\begin{table}[H]
\small
\setstretch{1.0}
\vspace{2mm}
\centering
\caption{Parameters used for evaluation.}
\label{tab:appendix-parameters}
\begin{threeparttable}
\begin{tabular}{cSSl}
	\toprule
	Parameter & {Value} & {Unit} \\ 
	\midrule
	$\costScalingFactor_{\text{ref}}$ & 1 & {-} \\
	$\costScalingFactor_{\text{lon}}$ & 35 & {-} \\
	$\costScalingFactor_{\text{lat}}$ & 50 & {-} \\
	\midrule
	$\samplingInterval$ & 1 & \si{\second} \\
	$\decelerationMax$ & -3 & \si{\metre\per\second\squared} \\
	$\sigma_a$ & 3 & \si{\metre\per\second\squared}  \\
	$t_{c, \min}$ & 1 & \si{\second} \\
	$t_{c, \max}$ & 5 & \si{\second} \\
	$l_{\text{goal}}$ & 15 & \si{\metre} \\
	$l_{\text{veh}}$ & 2 & \si{\metre} \\
	\midrule
	$l_{\text{idm}}$ & 2 & \si{\metre}  \\
	$T_{\text{idm}}$ & 1.5 & \si{\second} \\
	$\accelerationComfort$ & 0.73 & \si{\metre\per\second\squared} \\
	$\decelerationComfort$ & 1.67 & \si{\metre\per\second\squared} \\
	\midrule
	$\sigma_{\observation, \text{pos}}$ & 0.2 & \si{\metre} \\
	$\sigma_{\observation, \text{vel}}$ & 1.0 & \si{\metre\per\second} \\
	$d_{\text{thresh}}$ & 1 & \si{\metre} \\
	\midrule
	$\costScalingFactor_{\text{coll}}$ & -10000 & {-} \\
	$\costScalingFactor_{v}$ & -100 & {-} \\
	$\costScalingFactor_{j, \text{lon}}$ & -100 & {-} \\
	$\costScalingFactor_{a, \text{lat}}$ & -100 & {-}  \\
	\midrule
	$\omega$ & 0.77 & {-} \\
	$\discountFactor$ & 0.95 & {-} \\
	$\,\,\explorationConstant^{\dagger} $ & 10000 & {-} \\
	$\lipschitzConstant$ & 2000 & \si{\second\squared\per\metre} \\
	$d_{\text{roll}, \max}$ & 20 & {-} \\
	\bottomrule
\end{tabular}
\begin{tablenotes}[para,flushleft,small]
\footnotesize{\item[$\dagger$] We pick the value of exploration constant $\explorationConstant$ such that it meets the worst possible reward.}
\end{tablenotes}
\end{threeparttable}
\end{table}
\bibliographystyle{IEEEtran/IEEEtran}
\small{\bibliography{refs}}
\end{document}